\documentclass{article}
\usepackage{spconf,amsmath,graphicx,hyperref}
\usepackage{booktabs}
\usepackage{multirow}
\usepackage{adjustbox}
\usepackage{xcolor}
\usepackage{amsfonts}   
\usepackage{amssymb}
\usepackage{subcaption}

\definecolor{ForestGreen}{RGB}{0,100,0}

\title{Adapter-state Sharing CLIP for Parameter-efficient Multimodal Sarcasm Detection}

\name{Soumyadeep Jana \qquad Sahil Danayak \qquad Sanasam Ranbir Singh}

\address{Department of Computer Science and Engineering, 
	Indian Institute of Technology Guwahati, 
	India}

\begin{document}
%
\maketitle
\begin{abstract}
The growing prevalence of multimodal image-text sarcasm on social media poses challenges for opinion mining systems. Existing approaches rely on full fine-tuning of large models, making them unsuitable to adapt under resource-constrained settings. While recent parameter-efficient fine-tuning (PEFT) methods offer promise, their off-the-shelf use underperforms on complex tasks like sarcasm detection. We propose AdS-CLIP (Adapter-state Sharing in CLIP), a lightweight framework built on CLIP that inserts adapters only in the upper layers to preserve low-level unimodal representations in the lower layers and introduces a novel adapter-state sharing mechanism, where textual adapters guide visual ones to promote efficient cross-modal learning in the upper layers. Experiments on two public benchmarks demonstrate that AdS-CLIP not only outperforms standard PEFT methods but also existing multimodal baselines with significantly fewer trainable parameters.
\end{abstract}

\section{Introduction}
\label{sec:intro}
\begin{figure}[t]
    \centering
    \begin{tabular}{p{0.45\linewidth}p{0.45\linewidth}}
        \includegraphics[width=0.65\linewidth, height=0.65\linewidth]{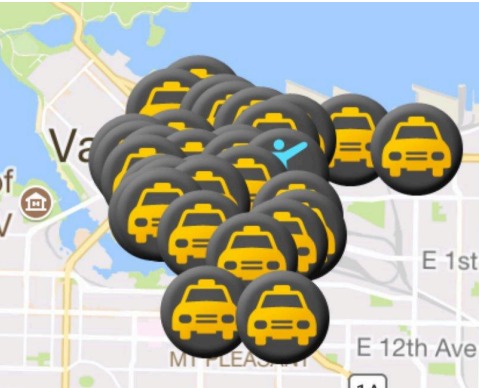} & 
        \includegraphics[width=0.65\linewidth, height=0.65\linewidth]{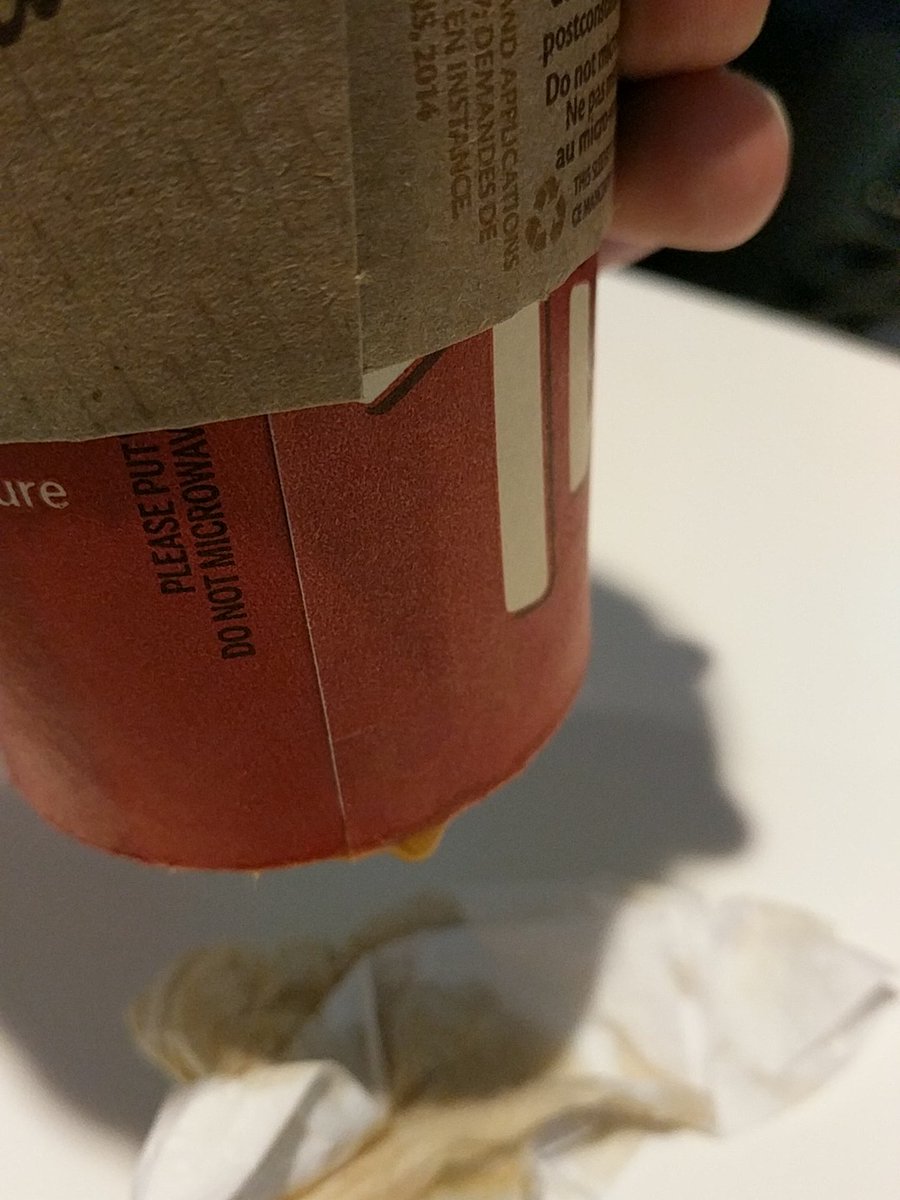} \\
        \footnotesize{(a) Ahh, such a shortage of cabs in the city.} &  
        \footnotesize{(b) Thanks for the awesome, leaky cup ... making my morning just so much better.} \\
    \end{tabular}
    \caption{Examples of multimodal sarcasm.} 
    \label{fig:multimodal_sarcasm}
\end{figure}

Sarcasm is a complex linguistic phenomenon in which the intended meaning of an utterance diverges from, and often opposes, its literal interpretation. On social media platforms, sarcasm frequently serves as a tool to mock, criticize, or express discontent. With the proliferation of multimodal content, the use of image–text sarcasm (Fig. \ref{fig:multimodal_sarcasm}) has grown rapidly, where incongruity between visual and textual modalities conveys sarcastic intent. The increasing prevalence of multimodal sarcasm presents significant challenges for opinion mining and sentiment analysis systems \cite{maynard-greenwood-2014-cares, badlani-etal-2019-ensemble, ghosh-etal-2021-laughing}.

Research on multimodal sarcasm detection initially focused on handcrafted fusion of image and text features \cite{Schifanella2016DetectingSI}. Later approaches used hierarchical fusion of image and text modalities \cite{cai-etal-2019-multi}, cross-modal semantic reasoning \cite{xu-etal-2020-reasoning}, and attention-based architectures to capture both intra- and inter-modal incongruities \cite{pan-etal-2020-modeling}. Graph-based techniques were subsequently proposed to model fine-grained token-level interactions between image and text tokens \cite{Liang2021MultiModalSD, Liang2022MultiModalSD}, while the integration of external knowledge further strengthened model performance \cite{liu-etal-2022-towards-multi-modal}. Recently \cite{qin-etal-2023-mmsd2} employed CLIP to detect sarcasm from image, text, and the combination of both views.

\begin{figure}[t]
\centering
\includegraphics[width=0.35\textwidth, height=0.2\textwidth]{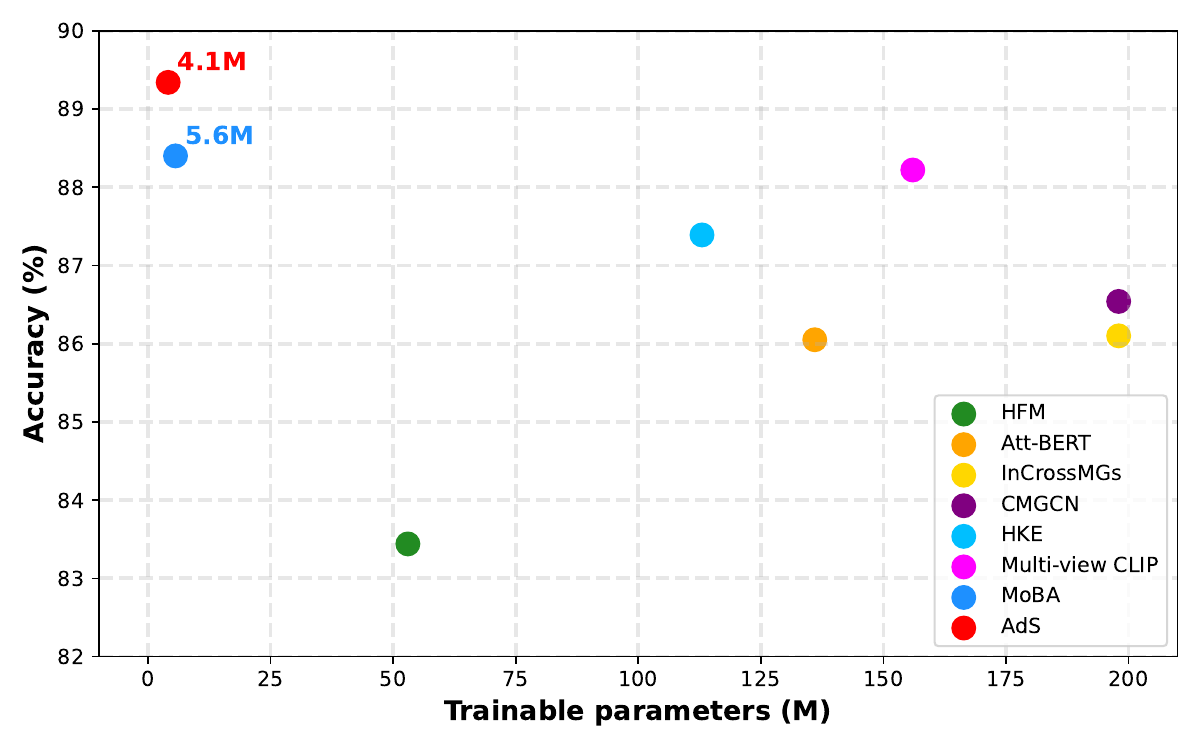} 
\caption{\label{fig:peft}Comparison of AdS-CLIP (red) vs other multimodal sarcasm baselines.}
\end{figure}

 Although prior approaches effectively modeled image–text incongruity, they relied on large number of trainable parameters, making them computationally expensive and less practical for finetuning in resource-constrained settings. To address parameter-efficieny, recently PEFT (Parameter-Efficient Fine-Tuning) methods such as adapter \cite{Houlsby2019ParameterEfficientTL}, prompt tuning \cite{Lester2021ThePO}, and LoRA \cite{Hu2021LoRALA} have become popular, which add only a small set of trainable parameters while keeping the backbone model frozen. However, direct adaptation of these methods yield suboptimal results for complex tasks like sarcasm detection (discussed in \S \ref{subsec:peft_comparison}). Building on this motivation, \cite{Xie2024MoBAMO} introduced MoBA for parameter-efficient multimodal sarcasm detection. Their approach stacks additional transformer layers over the CLIP encoders and inserts MoBA blocks between them. These blocks fuse the image and text signals using multiple adapters. While effective, their approach has two key limitations: (1) \textbf{Lack of cross-modal guidance:} In the MoBA block, both modalities are fused symmetrically, using multiple adapters, preventing the dominant text modality from guiding the image modality, which is essential for detecting sarcasm. 2) \textbf{Suboptimal placement:} By placing MoBA blocks between additional layers instead of within CLIP’s encoders, MoBA misses the chance to tightly integrate pretrained knowledge with task-specific adaptation, limiting its ability to capture sarcasm-relevant cues.

\indent To address these issues, we propose \textbf{AdS-CLIP} (\textbf{Ad}apter-state \textbf{S}haring in \textbf{CLIP}), a lightweight CLIP adaptation that leverages CLIP’s multimodal knowledge more effectively. To overcome the issue of \emph{suboptimal placement}, we insert adapters directly into CLIP’s existing encoders, avoiding extra layers and achieving tighter integration with pretrained features. An explorability study (discussed in \S \ref{subsec:adapter_placement}) further revealed that sarcasm is best captured in higher layers, motivating us to place adapters only in the top layers to learn abstract, sarcasm-relevant representations while preserving low-level features. To mitigate the \emph{lack of cross-modal guidance}, we introduce an adapter-state sharing mechanism, where textual adapters guide visual adapters, fostering directed cross-modal interaction. Experiments on two public datasets show that AdS-CLIP consistently outperforms baselines (\textcolor{ForestGreen}{+0.94 Acc / +0.39 F1} on MMSD and \textcolor{ForestGreen}{+0.42 Acc / +1.3 F1} on MMSD2.0) with only 4.1M trainable parameters (Fig.~\ref{fig:peft}). Additionally it also acheives superior performance over other PEFT methods (\textcolor{ForestGreen}{+1.0 Acc / +1.01 F1} on MMSD and \textcolor{ForestGreen}{+0.54 Acc / +0.58 F1} on MMSD2.0).

\textbf{Our contributions are three-fold:} (1) a novel adapter-state sharing mechanism for cross-modal interaction, (2) a selective top-layer adapter placement strategy that is more parameter-efficient than all-layer designs, and (3) a lightweight framework that achieves SOTA performance over existing baselines and PEFT methods.


\begin{figure}[t]
    \centering
    \includegraphics[width=\columnwidth, height=0.8\linewidth]{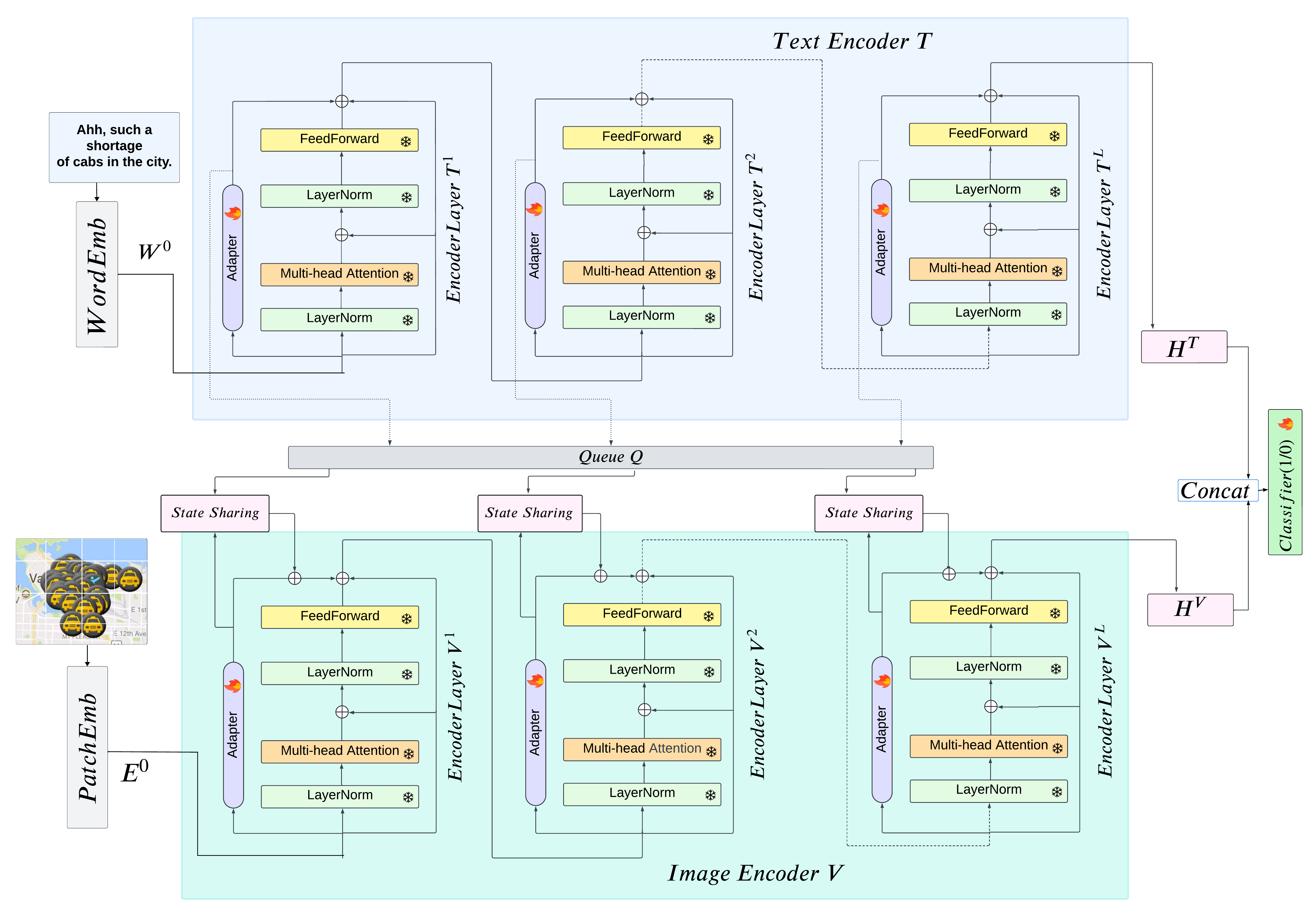}
    \caption{Architecture of AdS-CLIP model.}
    \label{fig:model}
\end{figure}

 \section{Preliminaries}
\textbf{CLIP Vision Encoder:} Image $I$ is divided into $m$ equal-sized patches. Each patch is projected into $d_{v}$ dimensional embedding space with positional embeddings to get the patch embedding $E^{0} (E^{0} \in \mathbb{R}^{m \times d_{v}} )$ for first transformer layer $V^{1}$. The patch embeddings $E^{i}$ are prepended with a learnable class embedding token $z_{i}$ and passed through L transformer blocks sequentially as follows:
\begin{equation}
[z^{i}, E^{i}] = \text{V}^{i}([z^{i-1}, E^{i-1}])
\quad i = 1, \dots, L
\end{equation}
The final image embedding $H^{V}$ is obtained by projecting $z^{L}$ into a shared embedding space as follows:
\begin{equation}
H^{V} = \text{Proj}_{I}(z^{L})
\quad H^{V} \in \mathbb{R}^{d}
\end{equation}
\textbf{CLIP Text Encoder:} Given an input text sequence $l$, it is tokenized into n tokens and are converted to $d_{t}$ dimensional word embeddings $W^{0} (W^{0} \in \mathbb{R}^{n \times d_{t}})$ with positional embeddings added.
Subsequently, they are processed sequentially through L transformer blocks as follows:
\begin{equation}
[W^{i}] = \text{T}^{i}([W^{i-1}])
\quad i = 1, \dots, L
\end{equation}
The final text embedding $H^{T}$ is obtained by projecting the embedding corresponding to the last token of the output from the last transformer layer into a shared embedding space as:
\begin{equation}
H^{T} = \text{Proj}_{T}(W^{L}[-1])
\quad H^{T} \in \mathbb{R}^{d}
\end{equation}

\section{Proposed Method}

\subsection{Problem Formulation}
Given a sample $x_{j} = (T_{j}, I_{j})$ where $T$ is the text and $I$ is the image, the goal is to assign $x_{j}$ with label ${y_{j}}$ from the set $Y = \{1, 0 \}$, where $1$ denotes \textit{sarcastic}.


\subsection{AdS-CLIP}
Figure \ref{fig:model} depicts the architecture of our proposed AdS-CLIP (Adapter-state Sharing in CLIP), which is a light-weight adaptation of CLIP. To adapt CLIP for our task in a parameter-efficient way, we incorporate adapters within both the textual and visual transformer blocks. Adapter $A$ is realized as $A(h) = W^{up}(\phi(W^{down}h))$ , where $W^{up}$ and $W^{down}$ are upsample and downsample projection layers respectively and $\phi$ is ReLU non-linearity.
Upper layers of CLIP learn more complex and abstract features \cite{Gandelsman2023InterpretingCI}. Since sarcasm is a complex phenomenon, we insert adapters into the upper transformer layers, ensuring that the model captures nuanced sarcasm cues while retaining CLIP’s foundational knowledge in the lower layers.
The text encoder can now be re-formulated with adapters added from $K^{th}$ transformer layer as:
\begin{equation}
\begin{split}
[W^{i}] & = \text{T}^{i}([W^{i-1}]) \quad i = 1, \dots, K-1 \\
[W^{i}] & = \text{T}^{i}([W^{i-1}]) + \alpha A_{T}^{i}([W^{i-1}])
\quad i = K, \dots, L
\end{split}
\end{equation}
where $ A_{T}^{i}$ is the adapter layer in the $i^{th}$ transformer block of the text encoder and $\alpha$ determines the retention factor of the adapter's information. 

Sarcasm relies heavily on image-text relationship. To effectively foster cross-modal information sharing, we introduce a novel adapter-state sharing mechanism, where the textual adapter states guide the visual adapter states to learn sarcastic cues from the image. To achieve this, we maintain a queue of textual adapter states $Q = \{ S_{j}^{T} \}_{j=K}^{L}$, where $S_{j}^{T} = A_{T}^{j}([W^{j-1}]) $  is the adapter output state from text transformer block $j$. For each visual adapter output, $S_{j}^{V} =  A_{V}^{j}([z^{j-1}, E^{j-1}])$, we compute its relative attention  weights corresponding to all textual adapter states using the following equations: $Att = softmax(QWS_{j}^{V})$ where $Q \in \mathbb{R}^{(L-K+1) \times d^{\prime}}$, $\quad W \in  \mathbb{R}^{d^{\prime} \times d},
\quad S_{j}^{V} \in  \mathbb{R}^{d \times 1}$, and $Att \in \mathbb{R}^{(L-K+1) \times 1}$.


$Att$ gives us the relative attention weights of each visual adapter state $S_{j}^{V}$ with all textual adapter states. To derive the final contribution $f^{j}$ of all textual adapter states, we perform a weighted summation using the computed attention scores.
\begin{equation}
   f^{j} = \sum\limits_{i=1}^{L-K+1}Att_{i} \cdot Q_{i}
   \quad f^{j} \in \mathbb{R}^{d}
\end{equation}
We now introduce adapters in the visual encoder from the $K^{th}$ transformer layer as:
\vspace{-0.05cm}
\begin{equation}
\begin{split}
[z^{j}, E^{j}] &= \text{V}^{j}([z^{j-1}, E^{j-1}])
\quad j = 1, \dots, K-1 \\
[z^{j}, E^{j}] &= \text{V}^{j}([z^{j-1}, E^{j-1}]) + \alpha A_{V}^{j}([z^{j-1}, E^{j-1}]) + \gamma f^{j}
\end{split}
\end{equation}
\vspace{-0.8cm}
\[
\quad j = K, \dots, L
\]
where $ A_{V}^{j}$ is the adapter layer in the $j^{th}$ transformer block of the visual encoder, $\alpha$ determines the retention factor of the adapter's information, $f^{j}$ is the weighted text  adapter shared information for $j^{th}$ visual adapter and $\gamma$ determines the retention factor for $f^{j}$.

\begingroup
\renewcommand{\arraystretch}{0.99} 
\setlength{\tabcolsep}{6pt} 
\small
\begin{table}[t]
  \centering
  \caption{Statistics of datasets.}
  \label{tab:stats}
  \begin{adjustbox}{width=0.8\linewidth}
    \begin{tabular}{cccc}
      \toprule
      MMSD / MMSD2.0 & Train & Val & Test \\
      \midrule
      Sentences & 19816 / 19816 & 2410 / 2410 & 2409 / 2409 \\
      Positive  & 8642 / 9576   & 959 / 1042  & 959 / 1037  \\
      Negative  & 11174 / 10240 & 1451 / 1368 & 1450 / 1372 \\
      \bottomrule
    \end{tabular}
  \end{adjustbox}
\end{table}
\endgroup

\section{Model Training and Prediction}
We feed the concatenated representation [$H^T; H^V$] to a linear classifier $clf$ for classification. During training, we freeze all the transformer layers of CLIP's text and visual encoders. The text adapter layers $A_{T}^{K \; \text{to} \; L}$, the visual adapter layers $A_{V}^{K \; \text{to} \; L}$, and the classifier $clf$ are trained with cross entropy loss.

\section{Datasets and Experimental Settings}

We evaluate AdS-CLIP on the only two public image-text sarcasm datasets, MMSD \cite{cai-etal-2019-multi} and MMSD2.0 \cite{qin-etal-2023-mmsd2}, statistics are shown in Table \ref{tab:stats}. We use ViT-B/16 CLIP with a batch size of 16, learning rate of 0.0015, and SGD optimizer. $\gamma$ and $\alpha$ are set to 0.5 and 0.05, respectively. All hyperparameters are chosen empirically. For both datasets, adapters are inserted into layers 7–12 of both encoders with adapter dimension 32. The model is trained for 20 epochs, and the best checkpoint is selected using validation accuracy. All experiments are performed on 1xNvidia RTX 5000 24GB GPU. We compare our model with SOTA multimodal baselines reported in \cite{Xie2024MoBAMO}.

\begingroup
\begin{table}[t]
    \centering
    \renewcommand{\arraystretch}{1.1}
    \setlength{\tabcolsep}{2pt}
    \caption{Main results across datasets. * denotes results from respective papers (no code available). Other baselines are reimplemented. "TP" denotes the number of trainable parameters. $^{\dagger}$ denotes significant improvement with $p<0.05$.}
    \label{tab:results}
    \begin{adjustbox}{max width=\columnwidth}
    \begin{tabular}{lcccc|cccc}
        \toprule
        \multirow{2}{*}{\textbf{Model}} & \multicolumn{4}{c|}{\textbf{MMSD}} & \multicolumn{4}{c}{\textbf{MMSD2.0}} \\
        & \textbf{Acc.} & \textbf{P} & \textbf{R} & \textbf{F1} & \textbf{Acc.} & \textbf{P} & \textbf{R} & \textbf{F1} \\
        \midrule

        HFM \cite{cai-etal-2019-multi} (TP=53M) & 83.44 & 76.57 & 84.15 & 80.18 & 70.57 & 64.84 & 69.05 & 66.88 \\
        Att-BERT \cite{pan-etal-2020-modeling} (TP=136M) & 86.05 & 80.87 & 85.08 & 82.92 & 80.03 & 76.28 & 77.82 & 77.04 \\
        InCrossMGs \cite{Liang2021MultiModalSD}* (TP=198M) & 86.10 & 81.38 & 84.36 & 82.84 & - & - & - & - \\
        CMGCN \cite{Liang2022MultiModalSD}* (TP=198M) & 86.54 & - & - & 82.73 & 79.83 & 75.82 & 78.01 & 76.90 \\
        HKE \cite{liu-etal-2022-towards-multi-modal} (TP=113M) & 87.36 & 81.40 & 86.48 & 84.09 & 78.50 & 73.48 & 71.07 & 72.25 \\
        MV-CLIP \cite{qin-etal-2023-mmsd2} (TP=156M) & 88.33 & 82.66 & 88.65 & 88.55 & 85.14 & 80.33 & 88.24 & 84.09 \\
        MoBA \cite{Xie2024MoBAMO} (TP=5.6M) & 88.40 & 82.04 & 88.31 & 84.85 & 85.22 & 79.82 & \textbf{88.29} & 84.11 \\
       
        \midrule
         AdS-CLIP (ours) (TP=4.1M) & \textbf{89.34} $^{\dagger}$ & \textbf{88.74}$^{\dagger}$ & \textbf{89.18}$^{\dagger}$ & \textbf{88.94}$^{\dagger}$ & \textbf{85.64}$^{\dagger}$ & \textbf{85.28}$^{\dagger}$ & 85.60 & \textbf{85.41}$^{\dagger}$ \\
        \bottomrule
    \end{tabular}
    \end{adjustbox}
\end{table}

\endgroup

\begingroup
\begin{table}[t]
\centering
\small
\renewcommand{\arraystretch}{0.95}
\setlength{\tabcolsep}{5pt}
\caption{Comparison with other PEFT methods in CLIP. PT is prompt tokens, and rank is LoRA rank. $^{\dagger}$ denotes significant improvement with $p<0.05$. M denotes million.}
\label{tab:ads_ablation}
\begin{adjustbox}{width=0.8\columnwidth}
\begin{tabular}{lcccccc}
\toprule
 & \multicolumn{2}{c}{\textbf{MMSD}} & \multicolumn{2}{c}{\textbf{MMSD2.0}} & \#Trainable\\
\cmidrule(lr){2-3} \cmidrule(lr){4-5}
\textbf{PEFT Methods} & \textbf{Acc} & \textbf{F1} & \textbf{Acc} & \textbf{F1} & \textbf{Params (M)} \\
\midrule
Adapters (1-12) & 88.26 & 87.13 & 83.96 & 84.21 & 6.93 \\
Adapters (7-12) & 88.34 & 87.93 & 84.61 & 84.47 & 4.176 \\
Prompt-Tuning (PT=4)  & 86.04 & 85.88 & 84.21 & 83.09 & 0.04 \\
Prompt-Tuning (PT=2)  & 86.32 & 86.17 & 84.32 & 83.11 & 0.02 \\
LoRA (rank 16) & 87.81 & 86.83 & 85.10 & 84.83 & 5.86 \\
LoRA (rank 8) & 86.21 & 85.13 & 84.81 & 84.19 & 2.93 \\
\midrule
 AdS-CLIP (ours) & \textbf{89.34}$^{\dagger}$ & \textbf{88.94}$^{\dagger}$ & \textbf{85.64}$^{\dagger}$ & \textbf{85.41}$^{\dagger}$ & 4.177 \\

\bottomrule
\end{tabular}
\end{adjustbox}
\end{table}
\endgroup

\section{Main Results}
We use accuracy (Acc), macro-average precision (P), macro-average recall (R), and macro-average F1 (F1) to report the performance in Table \ref{tab:results}. Our key observations are follows:
1) \textbf{SOTA results:} Despite having only 4.1M trainable parameters, AdS-CLIP surpasses all multimodal baselines that require hundreds of millions of parameters. This efficiency comes from minimally adapting CLIP’s rich pretrained features with lightweight adapters. (2) \textbf{Comparison with MoBA:} Compared to MoBA (5.6M params), AdS-CLIP achieves superior results with fewer params (4.1M). Specifically, it yields accuracy gains of $\delta$0.94\% on MMSD and $\delta$0.42\% on MMSD2.0, and F1 improvements of $\delta$0.39\% and $\delta$1.30\%, respectively. These gains highlight the benefits of optimal adapter placement and cross-modal state sharing.

\begingroup
\begin{table}[t]
\centering
\small
\renewcommand{\arraystretch}{0.9}
\setlength{\tabcolsep}{4pt}
\caption{Ablation of adapter placement for AdS-CLIP.}
\label{tab:adapter_placement}
\begin{adjustbox}{width=0.62\columnwidth}
\begin{tabular}{lcccccc}
\toprule
 & \multicolumn{2}{c}{\textbf{MMSD}} & \multicolumn{2}{c}{\textbf{MMSD2.0}} & \#Trainable\\
\cmidrule(lr){2-3} \cmidrule(lr){4-5}
\textbf{Layers} & \textbf{Acc} & \textbf{F1} & \textbf{Acc} & \textbf{F1} & \textbf{Params (M)} \\
\midrule
1-12   & 88.34 & 87.93 & 84.69 & 84.48 & 7.0 \\
3-12   & 89.05 & 88.58 & 84.60 & 84.36 & 6.0 \\
5-12   & 89.21 & 88.68 & 84.94 & 84.71 & 5.1 \\
7-12   & \textbf{89.34} & \textbf{88.94} & \textbf{85.64} & \textbf{85.41} & 4.1 \\
9-12   & 87.88 & 87.44 & 84.77 & 84.59 & 3.2 \\
11-12  & 87.55 & 87.06 & 85.31 & 85.02 & 2.2 \\
12     & 87.39 & 86.65 & 85.23 & 85.08 & 1.7 \\
\bottomrule
\end{tabular}
\end{adjustbox}
\end{table}
\endgroup

\subsection{Comparison with PEFT Methods}
\label{subsec:peft_comparison}


Table~\ref{tab:ads_ablation} compares AdS-CLIP with other CLIP-based PEFT techniques. Prompt tuning (adding learnable prompts to the embedding of all visual and text encoders) is highly parameter-efficient ($0.02$–$0.04$M), but its shallow adaptation at the input level limits deep cross-modal reasoning. Interestingly, increasing prompt length (PT=4) leads to lower performance, suggesting that longer prompts capture redundant information instead of improving task-specific cues. LoRA provides moderate gains, but higher-rank variants (rank 16) surpass AdS-CLIP in parameter cost but still underperforms. As observed in §7.1, placing adapters (even with no state sharing) in the upper layers (7–12) outperforms full-layer adapters with fewer parameters. AdS-CLIP builds on this with adapter-state sharing in these higher layers, delivering the best performance at nearly the same parameter cost.

\section{Analysis}
\subsection{Impact of Adapter Placement}
\label{subsec:adapter_placement}
Introducing adapters only in the upper layers of CLIP proves most effective for sarcasm detection. Prior work shows that while lower layers in CLIP capture modality-specific and local features, upper layers encode abstract, semantic representations crucial for complex reasoning tasks \cite{Radford2021LearningTV, Gandelsman2023InterpretingCI, Sammani2024InterpretingAA}. Since sarcasm hinges on such abstract incongruities, inserting adapters in the upper layers enables better task alignment. As shown in Table~\ref{tab:adapter_placement}, the best performance is achieved with adapters placed in layers 7–12, which also reduces trainable parameters compared to adding adapters in all layers.

\subsection{Embedding Space Visualization}
\begin{figure}[t]
    \centering
    \begin{subfigure}[t]{0.48\columnwidth}
        \centering
        \includegraphics[width=0.7\linewidth]{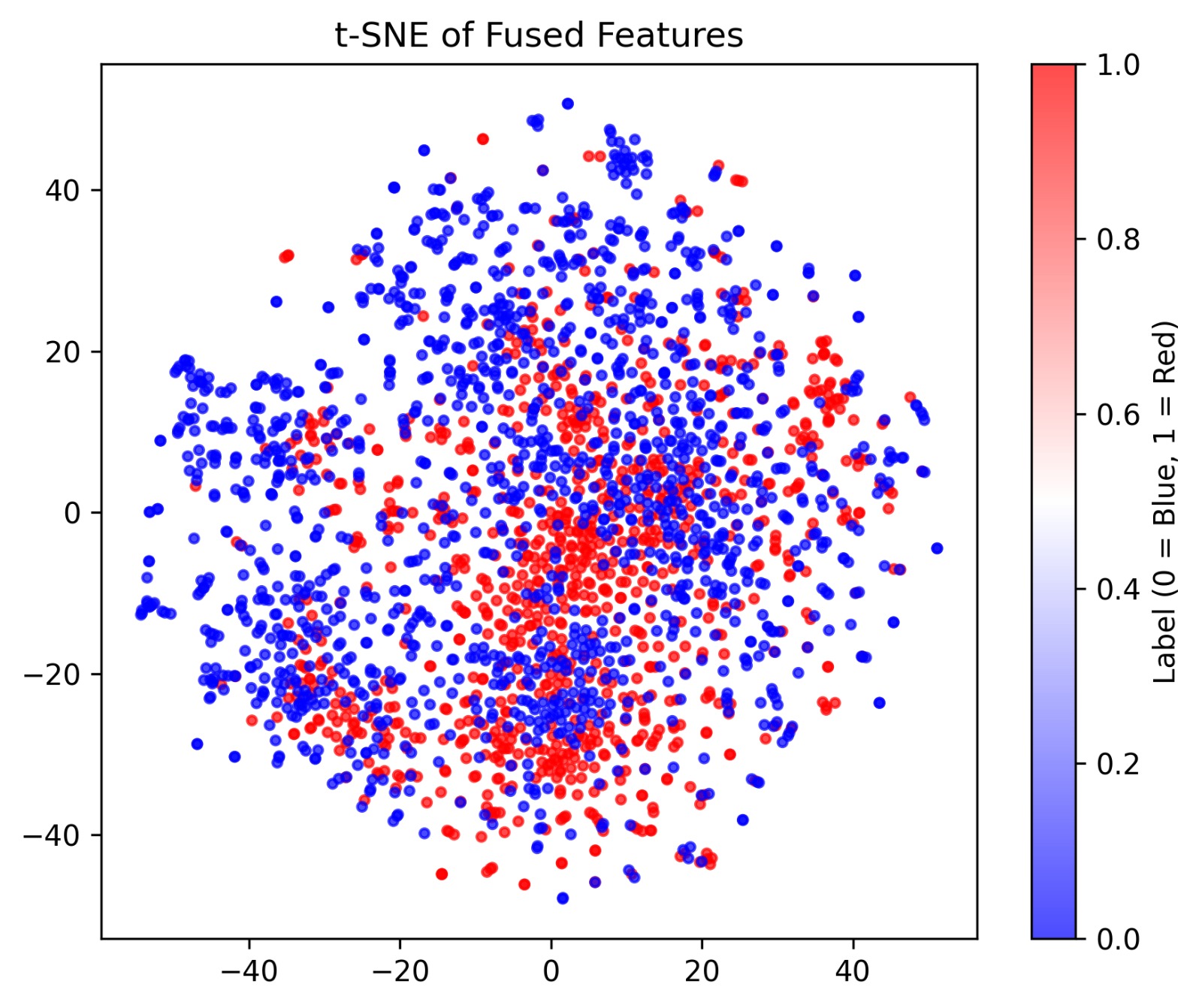}
        \caption{CLIP with Adapters (no state sharing)}
        \label{fig:vanilla-clip}
    \end{subfigure}
    \hfill
    \begin{subfigure}[t]{0.48\columnwidth}
        \centering
        \includegraphics[width=0.7\linewidth]{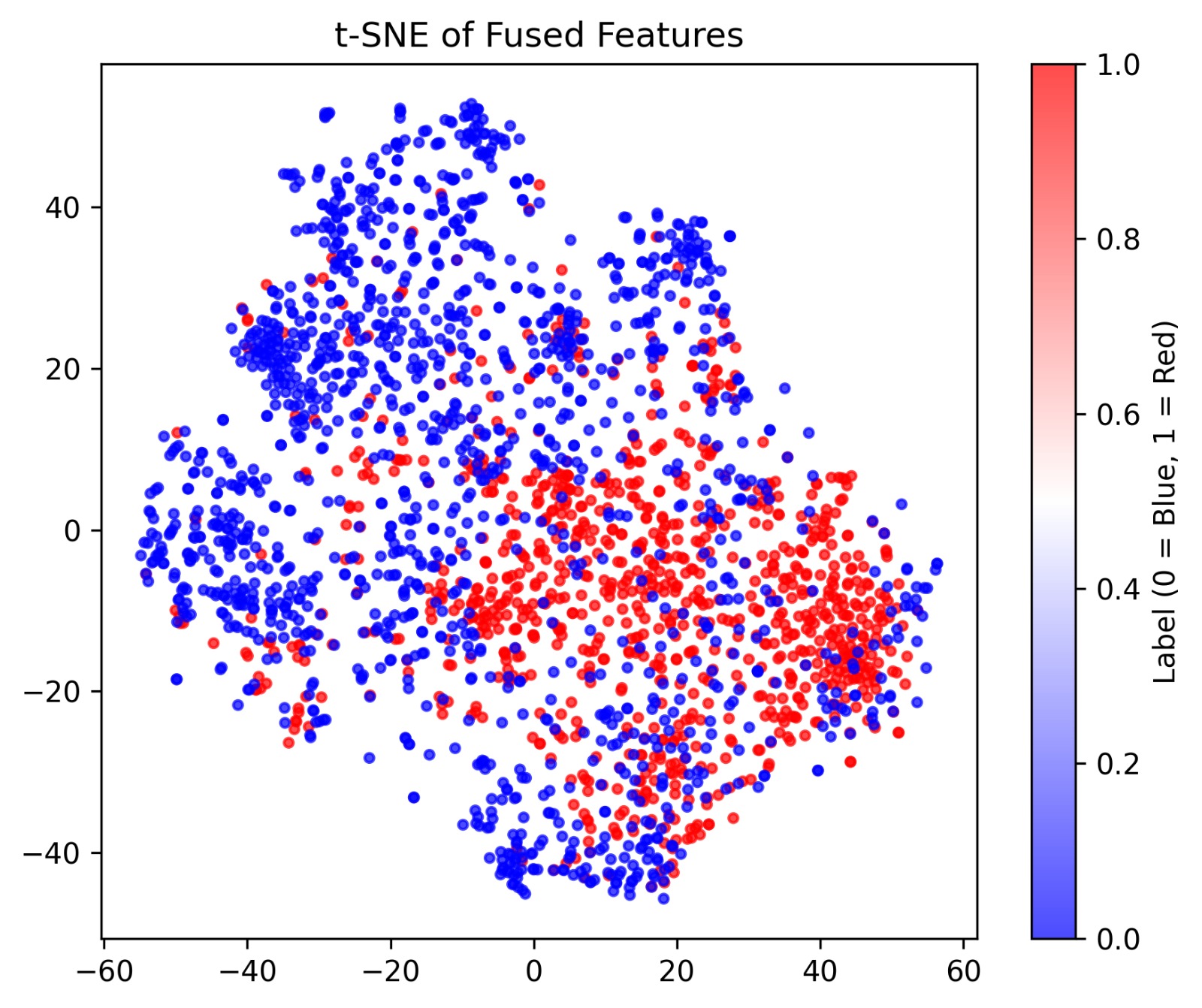}
        \caption{CLIP with AdS}
        \label{fig:ads-clip}
    \end{subfigure}
    \caption{Embedding space visualization}
    \label{fig:both}
\end{figure}

To better understand the impact of adapter-state sharing on a complex task like sarcasm detection, we visualize the concatenated embedding [$H^T; H^V$] before the classification layer using t-SNE.
When adapters are applied to the transformer layers of both visual and text encoders (Fig~\ref{fig:vanilla-clip}) (w/o state sharing), the embeddings are entangled, showing substantial overlap between both classes. In contrast, the AdS-CLIP model with adapter-state sharing (Fig~\ref{fig:ads-clip}) produces more structured and separable clusters, where sarcastic and non-sarcastic samples are better distinguished. 
This suggests that adapter-state sharing in higher layers enables the model to effectively capture task-specific nuances of sarcasm.

\begingroup

\begin{table}[t]
    \centering
    \renewcommand{\arraystretch}{0.98}

    \caption{Ablation of our model AdS-CLIP. w is with and w/o is without that specific component. $^{\dagger}$ denotes significant improvement with $p<0.05$.}
    \label{tab:ablation}
    \begin{adjustbox}{width=0.95\columnwidth}
    \begin{tabular}{lcccc|cccc}
        \toprule
        \multirow{2}{*}{\textbf{AdS}} & \multicolumn{4}{c|}{\textbf{MMSD}} & \multicolumn{4}{c}{\textbf{MMSD2.0}} \\
        & \textbf{Acc.} & \textbf{P} & \textbf{R} & \textbf{F1} & \textbf{Acc.} & \textbf{P} & \textbf{R} & \textbf{F1} \\
        \midrule
          w/o Adapters & 85.48 & 84.97 & 84.58 & 84.76 &  83.32 & 82.97 & 83.26 & 83.09\\
           w/o State Sharing & 88.34 & 87.68 & 88.27 & 87.93 &  84.61 & 84.28 & 84.47 & 84.37\\
          w V$\rightarrow$T State Sharing & 88.76 & 88.27 & 88.26 & 88.26 & 84.73 & 84.51 & 85.11 & 84.82\\
          \midrule
         AdS-CLIP (ours) & \textbf{89.34} $^{\dagger}$ & \textbf{88.74}$^{\dagger}$ & \textbf{89.18}$^{\dagger}$ & \textbf{88.94}$^{\dagger}$ & \textbf{85.64}$^{\dagger}$ & \textbf{85.28}$^{\dagger}$ & \textbf{85.60}$^{\dagger}$ & \textbf{85.41}$^{\dagger}$ \\
        \bottomrule
    \end{tabular}
    \end{adjustbox}
\end{table}
\endgroup

\section{Ablation}
To evaluate the strength of different components of our model, we design various ablation experiments. From Table \ref{tab:ablation}, we observe that the variant \textit{w/o Adapters} (both visual and textual adapters removed) shows reduced performance, highlighting the benefit of using the knowledge of CLIP through shared adapter learning. The variant \textit{$V\rightarrow T$ State Sharing} (visual adapters guide textual adapters) variant further degrades results, indicating that text adapters provides important cues to guide image adapters to attend to sarcasm specific regions. We also notice that \textit{w/o State Sharing} variant (adapters are independent) exhibits a drop in performance as the adapters lack cross-modal information sharing in this case.


\section{Conclusion}
In this work, we presented AdS-CLIP, a novel, parameter-efficient adapter-based adaptation of CLIP for multimodal sarcasm detection. By limiting adapter placement to the upper layers of CLIP and introducing a text-to-image adapter-state sharing mechanism, AdS-CLIP captures nuanced cross-modal sarcasm cues while maintaining low parameter overhead. Our method outperforms existing baselines and PEFT techniques, demonstrating that selective adapter placement and guided cross-modal interaction can significantly enhance performance with fewer parameter costs.

\vfill\pagebreak




\bibliographystyle{IEEEbib}
\bibliography{strings,refs}

\end{document}